\documentclass[11pt,a4paper]{scrartcl}

\usepackage{fontenc}
\usepackage[utf8]{inputenc}
\usepackage[english]{babel}

\usepackage[svgnames]{xcolor}

\usepackage{amssymb}
\usepackage{mathtools}
\usepackage{amsthm}
\usepackage[ruled, vlined, german, linesnumbered]{algorithm2e}
\usepackage{bm} %Für fettgedruckte Formeln (\boldsymbol{})
\usepackage{dsfont} %Für Indikatorfunktion-1
\usepackage{gauss}
\usepackage{enumitem} %für Listen mit variabler [leftmargin]

\usepackage{chngcntr}
\counterwithout{equation}{section}
\usepackage[title]{appendix}
\usepackage{hyperref}

\usepackage{tabularx}
\newcolumntype{L}[1]{>{\raggedright\arraybackslash}p{#1}} % linksbündig mit Breitenangabe
\newcolumntype{C}[1]{>{\centering\arraybackslash}p{#1}} % zentriert mit Breitenangabe
\newcolumntype{R}[1]{>{\raggedleft\arraybackslash}p{#1}} % rechtsbündig mit Breitenangabe

\usepackage{graphicx}
\usepackage{wrapfig} %Wrapfigures
\usepackage{caption} %Bildunterschriften
\usepackage{float} %für den Figure-Parameter [H] zur exakten Positionierung
\usepackage{tikz}
\usetikzlibrary{angles}
\usepackage{tkz-euclide}
\usepackage{pgfplots}
\usepgfplotslibrary{fillbetween}
\pgfplotsset{compat=1.11}
\tikzset{
  schraffiert/.style={pattern=horizontal lines,pattern color=#1},
  schraffiert/.default=black
}

\usepackage{geometry}
\newgeometry{left=2.5cm, right=2.5cm, top=3cm, bottom=4cm}
\usepackage{changepage} %für adjustwidth

\newcommand{\R}{\mathbb{R}}

\newcommand{\N}{\mathbb{N}}

\newcommand{\E}{\mathbb{E}}

\newcommand{\dd}{\:\text{d}}

\newcommand*{\vsep}{\kern-\tabcolsep\vrule height\arraystretch\ht\strutbox depth\arraystretch\dp\strutbox\kern-\tabcolsep} %Für vertikale Striche in Gauss-Matrizen

\allowdisplaybreaks

\usepackage{bold-extra}
\usepackage[dashed = false, style = authoryear, citestyle = numeric]{biblatex}

\makeatletter
\input{numeric.bbx}
\makeatother

\usepackage{xpatch}
\def\beginbold#1\endbold{\mkbibbold{#1}}
\xpretobibmacro{author}{\beginbold}{}{}
\xapptobibmacro{author}{\endbold}{}

\makeatletter
\patchcmd{\@maketitle}% <cmd>
  {{\usekomafont{date}{\@date \par}}%
    \vskip \z@ \@plus 1em}% <search>
  {}% <replace>
  {}{}% <success><failure>
\makeatother

\usepackage[plainheadsepline, headsepline]{scrlayer-scrpage}
\addtokomafont{section}{\large\mdseries\scshape}
\addtokomafont{subsection}{\large\mdseries\scshape}

\AtBeginDocument{
  \label{CorrectFirstPageLabel}
  
}

\begin{document}
\thispagestyle{empty}

\ihead{Interpolation of Missing Swaption Volatility Data using Variational Autoencoders}
\ohead{}

\vspace*{-1.2cm}
\noindent \rule{\textwidth}{3pt}
\vspace*{-0.45cm}
\hrule
\begin{center}
\noindent{\Large \textbf{Interpolation of Missing Swaption Volatility Data using \\ Gibbs Sampling on Variational Autoencoders}}
\end{center}
\vspace*{-0.4cm}
\noindent \rule{\textwidth}{1pt}
\begin{center}
\Large Ivo Richert and Robert Buch \\
\vspace*{0.2cm}
\large \today
\end{center}

\begin{center}
\normalsize \textsc{Abstract}
\end{center}
\vspace*{-0.7cm}
\begin{abstract} \footnotesize
\noindent Albeit of crucial interest for both financial practitioners and researchers, market-implied volatility data of European swaptions often exhibit large portions of missing quotes due to illiquidity of the various underlying swaption instruments. In this case, standard stochastic interpolation tools like the common SABR model often cannot be calibrated to observed implied volatility smiles, due to data being only available for the at-the-money quote of the respective underlying swaption. Here, we propose to infer the geometry of the full unknown implied volatility cube by learning stochastic latent representations of implied volatility cubes via variational autoencoders, enabling inference about the missing volatility data conditional on the observed data by an approximate Gibbs sampling approach. Imputed estimates of missing quotes can afterwards be used to fit a standard stochastic volatility model. Since training data for the employed variational autoencoder model is usually sparsely available, we test the robustness of the approach for a model trained on synthetic data on real market quotes and we show that SABR interpolated volatilites calibrated to reconstructed volatility cubes with artificially imputed missing values differ by not much more than two basis points compared to SABR fits calibrated to the complete cube. Moreover, we show how the imputation can be used to successfully set up delta-neutral portfolios for hedging purposes. 
\end{abstract}

\section{Introduction and related work} 

\noindent A complete interest rate swaption volatility cube is of practical interest for both researchers and practitioners, enabling inference about the current state of the market by the former, while making both hedging and consistent valuation of swaptions and more exotic derivatives through all maturities and tenors possible for the latter (\textcite{DK}).
The classic approach consists in fitting a given parametric model like the popular SABR model (\textcite{HetAl}) or the LIBOR market model to market observed smiles, i.e. slices of the full volatility cube, and in interpolating the missing quotes by the calibrated models as well as extrapolating beyond them (see e.g. \textcite{BM} for a variety of interest rate models).

Although the swaption market is approximately an order of magnitude larger than the next biggest interest rate derivatives market being the cap/floor market, larger market volumes do not necessarily mean that volatility quotes are liquid in all parts of the swaption volatility cube. Indeed, one often observes that the at-the-money swaption market is very liquid, however, for various tenors and expiries, the away from-the-money quotes are missing or not at all reliable, especially when compared to corresponding cap/floor volatilities (\textcite[p. 2]{SG}). When, however, for a given expiry and tenor, one only observes a single quoted strike of the whole smile (typically the at-the-money point), the stochastic model based approach cannot be applied directly. One then often needs to resort to simple interpolation schemes in order to fill the volatility cube and hence enable an ordinary calibration of a stochastic model. Given the sparsity of values in the away-from-the-money strike area however, simple linear or even more advanced interpolation schemes often work exceptionally bad or even fail to work at all, due to no available simplex of data points surrounding a missing value on the cubic grid. Then, artificial extrapolation methods need to be considered for the boundaries of the cube. For a more detailed discussion of the limited usability of interpolation schemes for preprocessing volatility data with missing values see \textcite[p. 8]{DK}. 

Alternatively, \textcite{HK} suggest fitting the SABR parameters $\beta$, $\rho$ and $\nu$ to the observed cap volatility surfaces and regressing the parameter $\alpha$ on the quoted swaption volatilites, while \textcite{JR} develop an explicit relationship between cap/floor and swaption volatilities by expressing the forward swap rate as a series of forward rates. These methods are referred to as ``lifting from caps'' and will not be considered further here. 

In contrast to the previously described methods for preprocessing volatility cubes with missing data for calibration of a stochastic model, we propose the idea of filling missing values in the volatility cube by a Gibbs sampling inspired approach proposed by \textcite{RMW} that is able to asymptotically sample from a learned variational approximation of the joint distribution of missing data and latent variables given the observed data of a volatility cube. Learning an approximative distribution driving the generation process of volatility cubes is carried out by the variational autoencoder (VAE) model of \textcite{KW} and \textcite{RMW}. Using this approach, missing values on the cubic grid can simply be reconstructed from the existing ones and can afterwards be used as an input in a standard SABR calibration procedure.

In addition to the Gibbs sampling-inspired approach described here, \textcite{SDetAl} and \textcite{GetAl} study imputation of missing data with a Markov chain whose transition operator is learned by an appropriate statistical model and with unsupervised clustering algorithms. Moreover, multiple other approaches for imputation of missing data were proposed by different authors. \textcite{BP} suggest using LSTM networks to solve a sequential Markov decision problem arising in the context of data imputation while \textcite{IFV} propose a variational autoencoder model that samples from a subset of missing features after being conditioned on arbitrary subsets of observed features.

Imputation of missing data using the Gibbs sampling-inspired approach of \textcite{RMW} was further studied in \textcite{MF1} and \textcite{MF2} who couple the algorithm described later with the importance-weighted autoencoder model from \textcite{BGS} in order to handle cases where training data of the employed deep latent variable models contain missing values. Furthermore, \textcite{MF1} propose to make use of a Metropolis-within-Gibbs extension of the algorithm presented here. However, we opt not to study this extension any further here, since the high dimensionality of the cubic swaption data makes calibrating a Metropolis proposal quite challenging.  

Since data for training the VAE model is sparsely available, and in particular not sufficiently available for training a machine learning model, we develop a method for generating synthetic swaption cubes from existing ones that can be used to train the VAE model. The robustness of this method is afterwards tested on real market-quoted volatility data and it is shown that SABR interpolated volatilites calibrated to reconstructed volatility cubes with artificially imputed missing values differ by not much more than two basis points compared to SABR fits calibrated to the true underlying complete swaption cube. 

The remaining paper is structured as follows. In the following section, we provide an overview over the variational inference paradigm utilized by the VAE model as well as over the Gibbs sampling-inspired approach of \textcite{RMW} in greater detail. The third section discusses means of synthetic data generation as it is necessary in the sparse swaption data environment. In the fourth section, we show the results of the Gibbs imputation on market-observed out-of-sample volatility cubes before we demonstrate how thereby obtained volatility values can be used to calibrate a SABR model to a volatility smile even when no quotes except from the at-the-money point are available and how one can successfully estimate the swaption's delta by the reconstructed volatility cubes. Lastly, the fifth section concludes.

\section{Variational Inference and Gibbs-Inspired Sampling}
The variational autoencoder model by \textcite{KW} aims to simultaneously train a generative model $p_\vartheta(x, z) = p_\vartheta(x | z) p(z)$ for the observable data $x \in \R^k$ given latent variables $z \in \R^d$ as well as an variational inference approximation $q_\theta(z | x)$ to the true posterior distribution $p(z|x)$. This is done by optimizing a variational lower bound to the logarithm of the intractable evidence $p_\vartheta(x) = \int p_\vartheta(x, z) \dd z$:
\begin{equation}\label{ELBO}
\log p_\vartheta (x) \geq \E_{q_\theta(z|x)} \left[ \log \frac{p_\vartheta (x, z)}{q_\theta (z | x)}\right] = \E_{q_\theta(z|x)} [\log p_\theta(x | z)] - \text{KL}(q_\theta(z|x) \lVert p(z))
\end{equation} 
where KL denotes the Kullback-Leibler divergence of two distributions. While the latent prior distribution is commonly modelled by a standard normal distribution, the likelihood $p_\vartheta(x | z)$ and the variational posterior $q_\theta(z | x)$ are assumed to be Gaussians with diagonal covariance matrix parameterized by neural networks, called encoder and decoder, with parameters $\theta$ and $\vartheta$ for which maximizing \eqref{ELBO} provides a tractable training criterion.\\
\\
After training a variational autoencoder model on a complete data sample, we now assume that the remaining data $x$ can be decomposed into an observed and into a missing component by $x = (x_{\text{obs}}, x_{\text{miss}})$. In order to infer the missing values inherent in the sample, the penultimate goal relies in sampling from the conditional distribution of $x_{\text{miss}}$ given $x_{\text{obs}}$. We now utilize a Gibbs sampling-inspired approach in order to sample from the conditional joint distribution of the random vector $(x_{\text{miss}}, z)$ given $x_{\text{obs}}$. Samples from the required conditional distribution of $x_{\text{miss}}$ given $x_{\text{obs}}$ are then obtained via marginalization of the samples from the distribution of  $(x_{\text{miss}}, z)$ given $x_{\text{obs}}$. To summarize, the algorithm for missing data imputation proceeds in the following way:

\begin{itemize}
\item[1)] Train a variational autoencoder to learn the encoder distribution $q_\theta(z|x)$ and the decoder distribution $p_\vartheta(x|z)$.
\item[2)] Choose starting values $x_{\text{miss}}^{(0)}$ and set $t=0$.
\item[3)] Simulate successively
\begin{align*}
z^{(t+1)} &\sim q_\theta(z | (x_{\text{obs}}, x_{\text{miss}}^{(t)})), \\
x^{(t+1)}_{\text{miss}} &\sim p_\vartheta(x_{\text{miss}} | x_{\text{obs}}, z^{(t+1)}).
\end{align*}
Here, $p_\vartheta(x_{\text{miss}} | x_{\text{obs}}, z^{(t+1)})$ denotes the conditional distribution of the missing data given the observed data and the current latent code which is obtained by conditioning $p_\vartheta$ on $x_{\text{obs}}$. In the following, as it is typical, the decoder $p_\vartheta$ is modelled by a multivariate Gaussian with diagonal covariance matrix, hence the conditioning on $x_{\text{obs}}$ can effectively be omitted. %Alternatively, the decoder covariance matrix can for example be modeled as a ``diagonal + rank-1'' matrix as in \textcite{RMW}.
\item[4)] Increment $t$ and return to step 3) until the maximum number $T$ of iterations is reached. Then go to step 5).
\item[5)] Obtain a sequence of marginals $(x^{(t)}_{\text{miss}})_{0 \leq t \leq T}$ by discarding $z^{(t)}$ from the obtained Markov chain of vectors $(x_{\text{miss}}, z^{(t)})_{t \in \N_0}$. 
\end{itemize}
Clearly, the process $(x^{(t)}_{\text{miss}})_{0 \leq t \leq T}$ generated by the above is a Markov chain. \textcite{RS} give conditions for aperiodicity and irreducibility of the Gibbs chain resulting in convergence to the stationary distribution. Moreover, under rather mild conditions, it can be shown (see \textcite{GG}) that the stationary distribution of the above Markov chain comes out as an approximation of the true conditional distribution of $x_{\text{miss}}$ given $x_{\text{obs}}$ which we call $$q(x_{\text{miss}} | x_{\text{obs}}) := \int p_\vartheta(x_{\text{miss}} | z, x_{\text{obs}}) q_\theta(z | x_{\text{obs}}) \dd z.$$ The above procedure is termed Pseudo-Gibbs sampling. This is due to the fact that in step 3) of the above algorithm we just use the variational posterior approximation $q_\theta$ instead of the true posterior $p(z|x)$. If $q_\theta(z|x)$ and $p(z|x)$ coincide, the above algorithm exactly coincides with the Gibbs sampling framework and we obtain samples from the true conditional distribution $$p(x_{\text{miss}} | x_{\text{obs}}) = \int p_\vartheta(x_{\text{miss}} | z, x_{\text{obs}}) p(z | x_{\text{obs}}) \dd z.$$ Using Birkhoff's ergodic theorem, we can use realizations from the chain to estimate the expectation of any integrable function of $x_{\text{miss}}$ conditional on $x_{\text{obs}}$ under $q$.

In practice, the first $M$ values of the chain are discarded as a ``burn-in'' where $M$ is chosen sufficiently large in order to ensure convergence of the chain to the stationary distribution. Finally, after obtaining the Markov chain $(x^{(t)}_{\text{miss}})_{M \leq t \leq T}$, we impute the missing values $x_{\text{miss}}$ of the data sample by an estimator of the conditional expectation of $x_{\text{miss}}$ given $x_{\text{obs}}$ under $q(x_{\text{miss}} | x_{\text{obs}})$ which, due to Birkhoff's ergodic theorem, is given by the sample average \vspace*{-0.3cm}\begin{equation}\label{avrg}
\hat{x}_{\text{miss}} := \hat{\E}_{q(x_{\text{miss}} | x_{\text{obs}})} (x_{\text{miss}} | x_{\text{obs}}) := \frac{1}{T - M + 1} \sum_{t=M}^T x^{(t)}_{\text{miss}}.
\end{equation}

\section{Employed Data Sets and Synthetic Data Generation}
In order to train the variational autoencoder model, daily (Bachelier model implied) swaption volatility cubes of European LIBOR swaptions were obtained from the FENICS market data provider between 29 September 2021 and 21 August 2019. Since as of now, due to missing data within this 2 year period, there are only around 120 full observed swaption volatilty cubes in our dataset, we need to apply synthetic data augmentation methods to provide enough training data coming from heterogeneous market states. Thereby, we enhance robustness and generalization capabilites while limiting overfitting in the VAE model. We decide to apply synthetic simulation in the parameter domain of the SABR stochastic volatility model (see Appendix A). This is similar to an approach described in \textcite{Adr}.

In order to obtain realistic synthetic data, we fit the SABR stochastic volatility model with a fixed $\beta = 0.5$ to the 120 swaption cubes in the dataset. As it is the market standard practice, we model the swap rate for each tenor $T_M$ and maturity $T_0$ (see Appendix A for further notation) independently by a SABR model with parameters $\alpha_{T_0, T_M}$, $\nu_{T_0, T_M}$ and $\rho_{T_0, T_M}$. This way, we obtain a dataset of matrices of parameter values for each day and each parameter.

Afterwards, we generate a synthetic $\alpha$-parameter matrix in the following way: First we generate the upper and left boundaries of the parameter matrix corresponding to the lowest swaption tenor and lowest swaption maturity by fitting normal distributions to the increments between adjacent parameter values in the boundaries and afterwards sampling from these increments. After obtaining the upper and left boundaries of the parameter matrix we consecutively generate the values of parameters in the rest of the coordinates of the parameter matrix in the following way: The $(i, j)$-th element $\alpha_{i, j}$ of the parameter matrix is obtained by fitting a normal distribution to the values $(\alpha_{i, j} - \alpha_{i-1, j}) / (\alpha_{i-1, j} - \alpha_{i, j-1})$ in the dataset which describe how the relative position $\alpha_{i, j}$ between the $(i-1, j)$-th and $(i, j-1)$-th value in the matrix is distributed and afterwards sampling from this distribution. This way, we maintain the inherent structure of the parameter values across the $(T_0, T_M)$-grid while simultaneously enriching the sparse dataset with new unseen parameter combinations.

\begin{figure}[H]
\centering
\includegraphics[trim = 0mm 0mm 0mm 5mm, clip, width=0.93\textwidth]{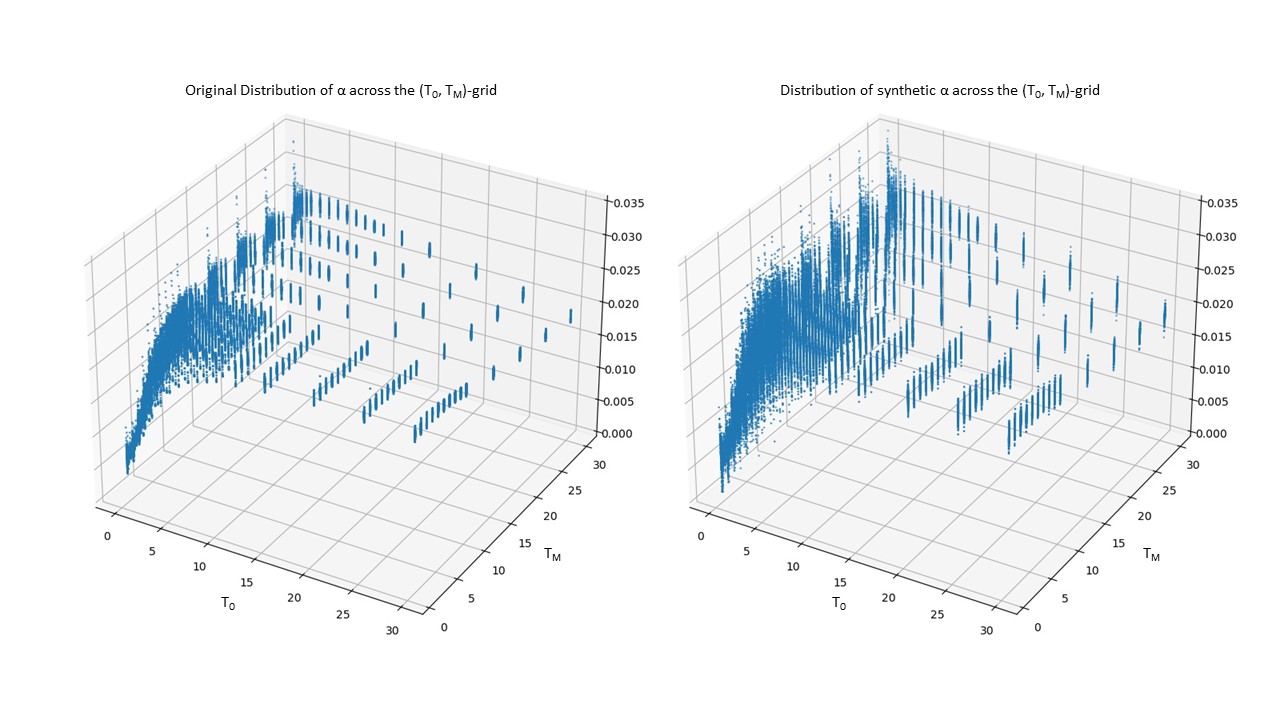}
\caption{Scatterplot of the original distribution of the $\alpha$-parameter across the $(T_0, T_M)$ grid on 20 October 2021 next to the distribution of synthetically samples $\alpha$-parameters. The synthetic distribution displays relative overdispersion compared to the original distribution, leading to a greater variety of volatility cubes for VAE training.}
\label{AlphaDist}
\end{figure}

A similar approach was used for generating synthetic matrices of the parameters $\nu$ and $\rho$. In order to make sure that the sampled $\alpha$-, $\nu$- and $\rho$-values are positive and $[-1, 1]$-supported respectively, we apply a log-transform to the $\alpha$- and $\nu$-parameter matrices and an artanh-transform (Fisher's $z$-transform) to the $\rho$-parameter matrices before sampling. A scatterplot of the original distribution of the $\alpha$-parameter across the $(T_0, T_M)$-grid next to the distribution of synthetically sampled $\alpha$-parameters can be found in Figure \ref{AlphaDist}. 

After obtaining synthetic realisations of SABR parameter matrices across the grid of swaption maturities and tenors, we transform them back into swaption volatility cubes using the implied volatility expansion formulas from \textcite{HetAl} (see Appendix A). 

We note here that, due to the random simulation practice, the obtained swaption cubes exhibit somewhat more roughness compared to the market observed ones. We opt not to smooth out the synthetic data before training in order to make use of the denoising capabilities of VAE. It has been extensively studied (see e.g. \textcite{VetAl}) that inducing additional noise to the input improves generalization performances of deterministic autoencoder models, since it enhances robustness of adjacent data points in the latent manifold against the
presence of small noise in the higher dimensional observation space. Moreover, \textcite{RMW} advocate that denoising enhances the generalization capabilities of probabilistic generative models as well, by noting that additional input noise is crucial for the recognition model to achieve the desired accuracy on unseen data.

\vspace*{-0.3cm}

\section{Empirical results using VAE-imputation on real market data}

\begin{figure}[H]
\centering
\includegraphics[width=0.8\textwidth]{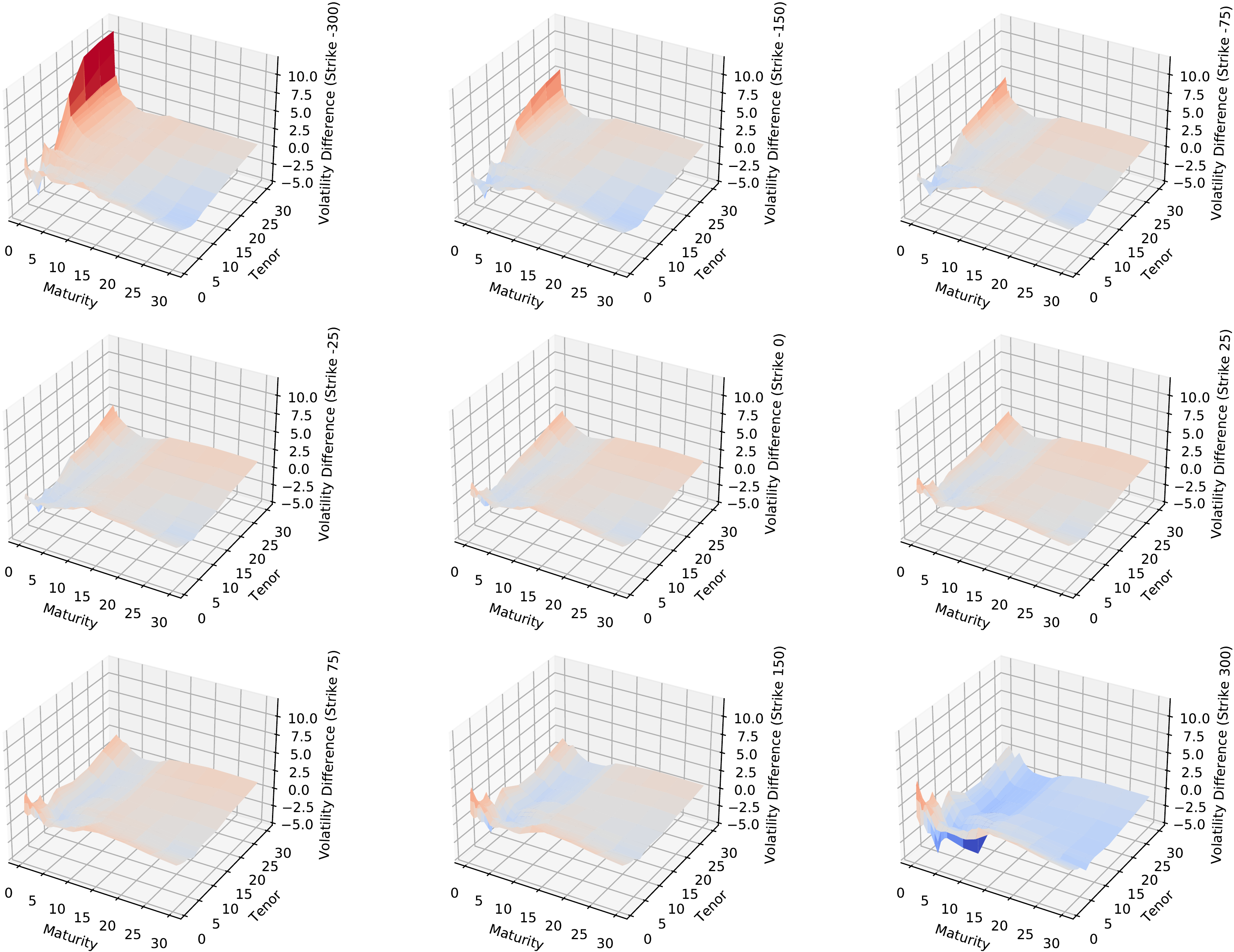}
\caption{Difference between the market observed swaption volatility cube on 20 October 2021 and a VAE reconstructed volatility cube after 79.6\% of the values were treated as missings.}
\label{VolDiff}
\end{figure}

\noindent Figure \ref{VolDiff} shows the difference between the basis point volatility cube (Bachelier model implied normal volatilities) of European LIBOR swaptions on 20 October 2021 and a reconstructed volatility cube using the VAE imputation from Section 2 which was obtained in the following way: Approximately 79.6\% of the observations on the cube were set zero and treated as missing data before the Pseudo-Gibbs imputation algorithm was run on the cube. The locations of the missing points were chosen to match the locations of missing points of the volatility cube on 21 August 2019, of which only roughly over 20\% of the values on the full three dimensional grid were not missing. The pretrained inference model was a standard variational autoencoder with a ten-dimensional latent space which was trained for 50000 epochs on 10000 synthetically generated swaption cubes by the procedure described in Section 2. The missing values were imputed by the sample averages \eqref{avrg} of the samples from a Gibbs Markov chain of length 2000 with a burn-in period of 100. More details about the VAE model architecture and the original and reconstructed cube can be found in Appendix C. The mean absolute deviation of the imputed values from the true values is 1.9123 basis points, demonstrating the good out-of-sample generalization capabilities of the approach. The algorithm exhibits its greatest uncertainty of reconstruction in the low maturity - high tenor region which is also resembled in Figure \ref{SABR_Maes}.

We note here that in our setup it is of no importance to make sure whether the imputed volatility cubes after Gibbs sampling exhibit arbitrage or not, since we opt to use the VAE imputed cube as a mere input for the calibration procedure of a subsequent stochastic volatility model like the SABR model. Maintaining no-arbitrage conditions then depends of the choice of the specific stochastic model employed, see e.g. \textcite{JN}.

\noindent Convergence of the imputed sample averages \eqref{avrg} can be monitored at each step by standard techniques for computation of confidence intervals from Markov chain Monte Carlo theory, e.g. the Overlapping Batch Means estimator of \textcite{FJ}. Note that a 95\%-confidence error of $\epsilon$ does not imply a 95\% chance of the interval $[\hat{x}_{\text{miss}} - \epsilon, \hat{x}_{\text{miss}} + \epsilon]$ to cover the masked value of the observed volatility cube that shall be imputed, but merely a 95\% chance of the interval to cover the value $\E_{q(x_{\text{miss}} | x_{\text{obs}})} (x_{\text{miss}} | x_{\text{obs}})$ whatever this value will be. Thus, we cannot expect the imputed cube to ``converge'' to the true observed cube but merely to an approximation of the cube that we would expect after seeing the observed values and given the underlying distribution of data that was inferred from our synthetic training data. 

\begin{figure}[H]
\centering
\includegraphics[width=0.92\textwidth]{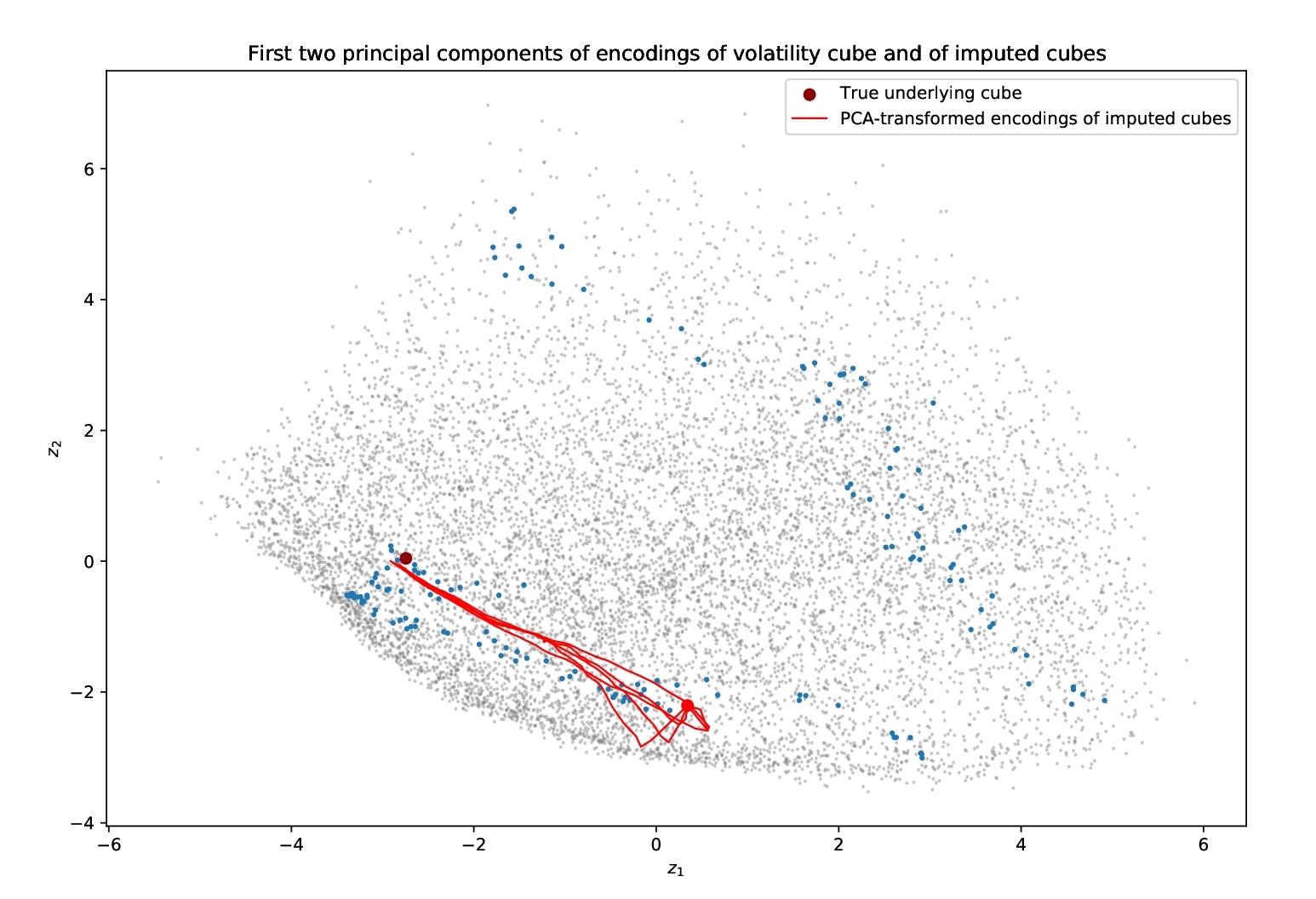}
\caption{First two principal components of ten-dimensional latent encodings of synthetic and market observed volatility cubes as well as of the cube on 20 October 2021 in dark red. Displayed are the means of the Gaussian VAE-approximated posteriors $q_\theta(z|x_i)$. Blue dots indicate encodings of market observed data while the grey dots show the encodings of the synthetic VAE training data. Furthermore, the paths that were traced by the latent encodings of the sample average imputed cubes after between 1 and 2000 Markov chain steps are shown in red for five different runs of the VAE imputation algorithm.}
\label{Encodings}
\end{figure}

Convergence can also be monitored in the latent space instead of in the observable space. Figure \ref{Encodings} displays the first two principal components of the ten-dimensional latent encodings of the synthetic and market observed volatility cubes as well as in dark red the latent encoding of the volatility cube from 20 October 2021 that was used for Figure \ref{VolDiff}. Furthermore, the paths that were traced by the latent encodings of the sample average imputed cubes between 1 and 2000 Markov chain steps are shown in red for five different runs of the Pseudo-Gibbs imputation algorithm. It is seen that the latent encodings of the sample averages converge to some encoding near the encoding of the true underlying cube.

Studying the underlying non-euclidean geometry of the latent space of the VAE model that is displayed in the above Figure \ref{Encodings} provides an interesting possibility for further reseach. Lately, it has been noted (see \textcite{AHH} and \textcite{H}) that, empirically, Euclidean latent space distances carry little information about the relationship between data points and that an interpretation of the latent space as a Riemannian manifold appears more promising. Hence, it could be fruitful to investigate whether the paths traced by the imputed cubes via the Pseudo-Gibbs algorithm in Figure \ref{Encodings} are in concordance with such a geometric structure, i.e. whether these paths represent geodesics on the respective manifolds between the initial latent representation of the missing cube and the latent representation of the true underlying cube. 

After obtaining an imputed reconstruction of a swaption volatility cube containing missing values, these reconstructions can be used to calibrate certain market-standard stochastic volatility models whose calibration would otherwise fail because of a too sparse data domain. We follow this procedure in the following section using the example of the SABR stochastic volatility model (see Appendix A).

\subsection{Fitting the SABR model to imputed volatility smiles}
\begin{figure}[H]
\centering
\includegraphics[trim = 0mm 0mm 0mm 5mm, clip, width=0.88\textwidth]{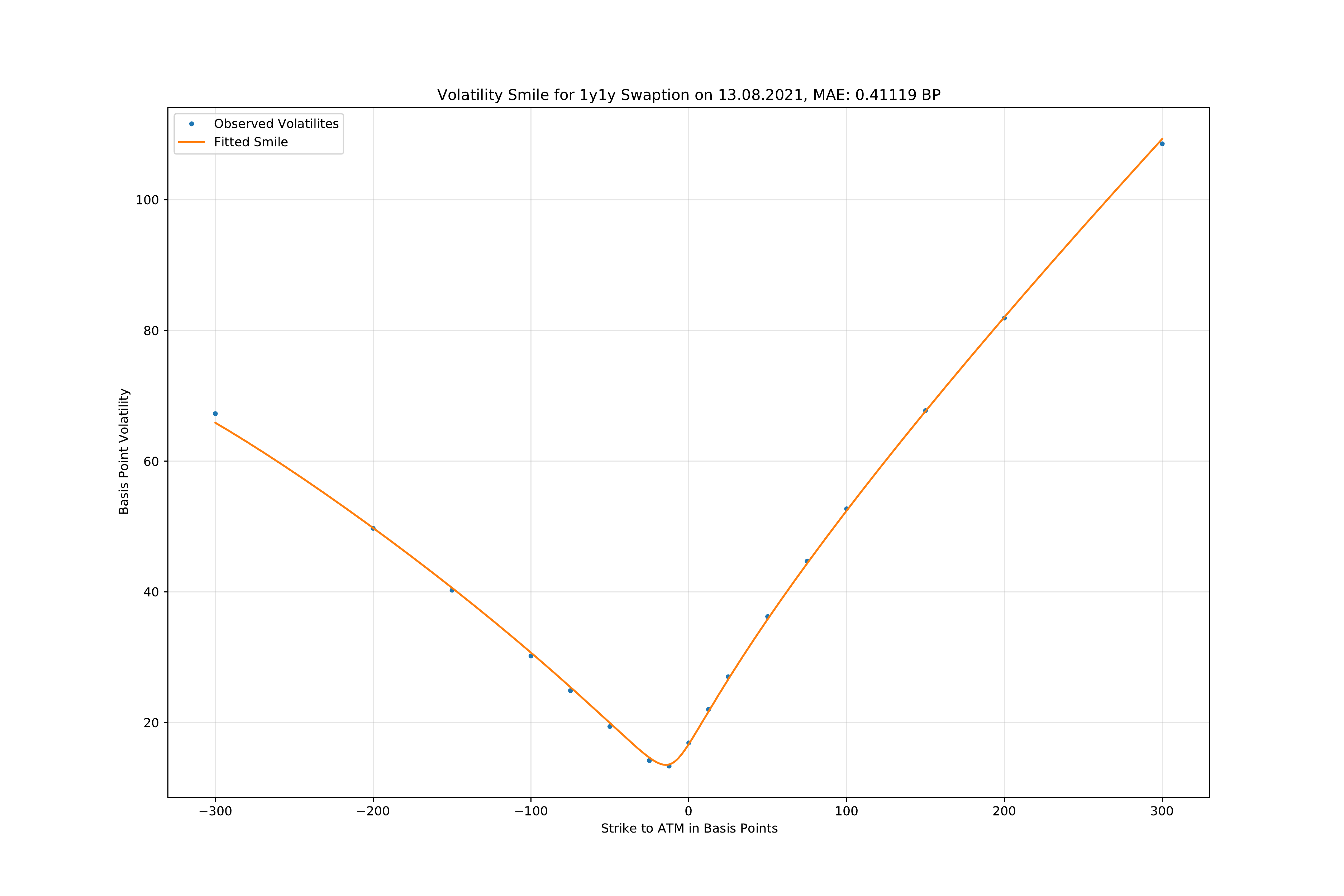}
\caption{SABR fitted volatility smile for the 1 year by 1 year swaption on 20 Oct 2021. The calibrated parameters are $\alpha=0.0086$, $\nu = 1.0732$ and $\rho= 0.6506$ and the mean absolute error of fit is 0.35 basis points.}
\label{SABR}
\end{figure}

In the following, for each tenor-maturity-combination, we fix $\beta$ to a value of 0.5 and fit the other three parameters of the SABR stochastic volatility model by a standard least-squares minimization. The shift introduced is $b=0.04$ or 400 basis points to ensure positivity of all forward swap rates on each date as well as of all strikes. Figure \ref{SABR} shows the calibrated smile the model produces for the swaption with maturity and tenor of 1 year on 20 October 2021.

\begin{figure}[H]
\centering
\includegraphics[trim = 0mm 0mm 0mm 5mm, clip, width=0.85\textwidth]{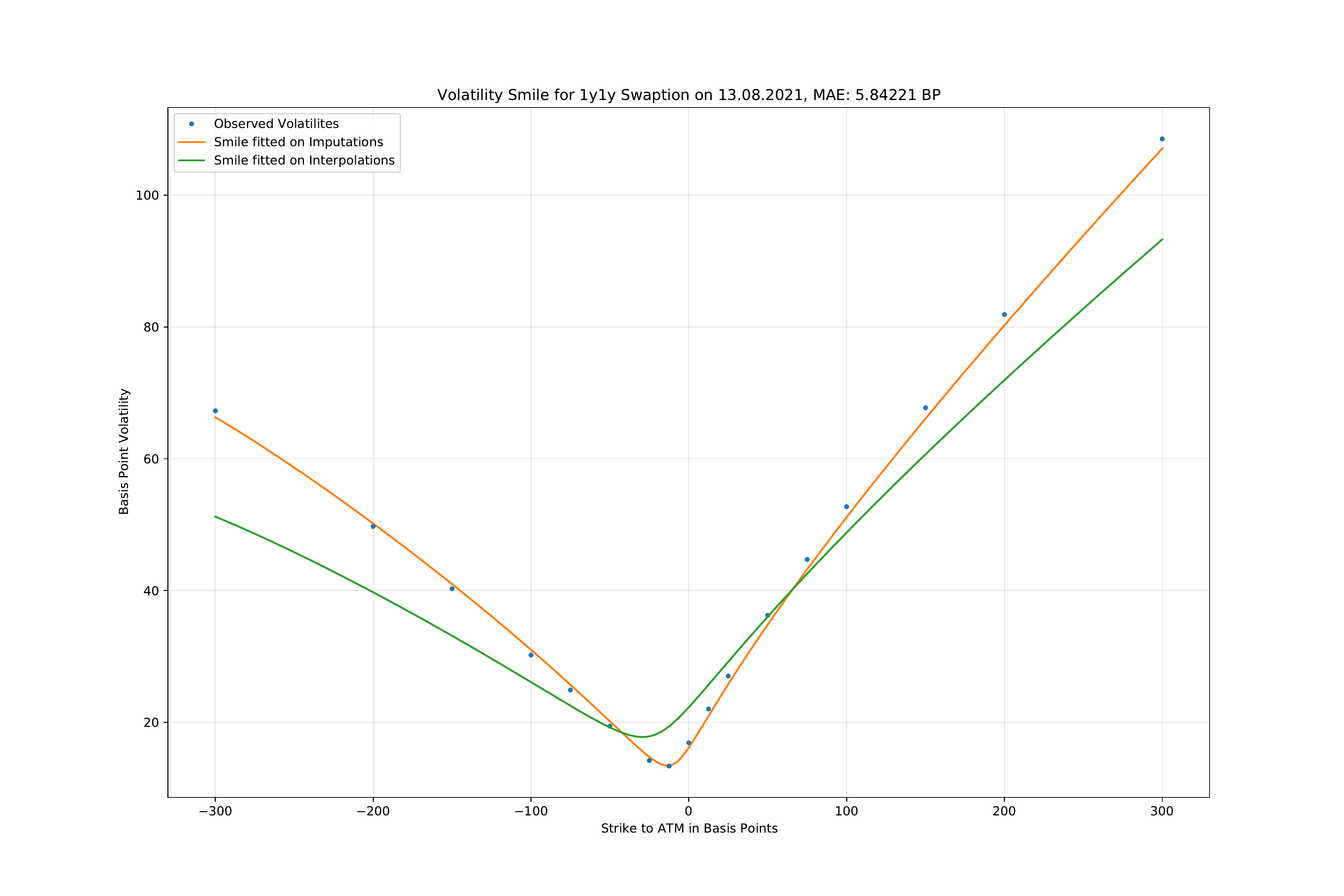}
\caption{SABR fitted volatility smile for the 1 year by 1 year swaption on 20 Oct 2021 when all but the at-the-money point are replaced by the imputed values from the Pseudo-Gibbs algorithm and by linearly interpolated values on the cube respectively. The calibrated parameters are $\alpha=0.0083$, $\nu = 1.0622$ and $\rho= 0.6169$ in the first case and $\alpha=0.0117$, $\nu = 0.7349$ and $\rho= 0.6561$ in the second case.}
\label{SABR_Reco}
\end{figure}

Since for this swaption, all volatility quotes except from the at-the-money point are missing in the volatility cube of between 21 Aug 2019 and 24 Apr 2020, the aim is now to examine how much the fit of the SABR model varies when the true volatility quotes except from the at-the-money point are replaced by ones from the Pseudo-Gibbs imputation and by ones from a simple interpolation. Below, Figure \ref{SABR_Reco} shows the calibrated SABR volatility smile for the same swaption on the same date, where except for the at-the-money point all other volatility quotes are replaced by a) the Pseudo-Gibbs imputed volatility values and b) volatility values from a simple interpolation (respectively extrapolation on the boundary of the observed market cube) of the swaption cube at the missing values.

The interpolated volatility cube clearly fails to adequately rebuild the true volatility structure due to the large amount of missing values which results in an insufficient fit of the smile and a mean absolute deviation from the true smile of around 5.84 basis points. This does not happen to the same extent with the Pseudo-Gibbs imputed volatility cube which exhibits only a mean absolute deviation of approximately 1.05 basis points from the true smile.

\begin{figure}[H]
\centering
\includegraphics[ width=0.76\textwidth]{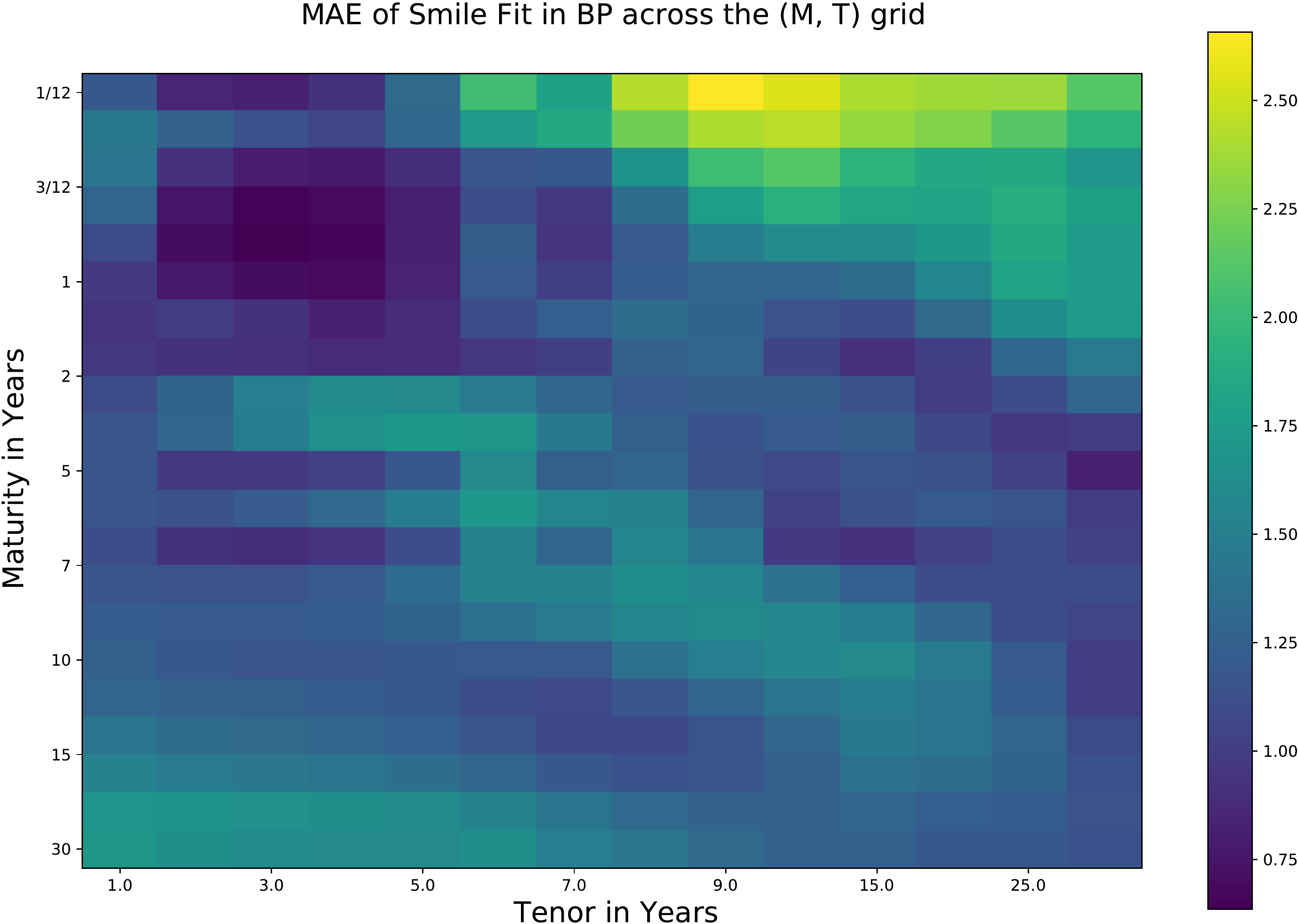}
\caption{Mean Absolute Errors between observed volatilities of different swaptions on 20 Oct 2021 and SABR fitted volatility using the imputed values from the Pseudo-Gibbs algorithm.}
\label{SABR_Maes}
\end{figure}

An analysis of the mean absolute error of the SABR fits calibrated on VAE-imputed volatilities in the same manner on all different swaptions on the $(T_0, T_M)$-grid can be found in Figure \ref{SABR_Maes}. The worst fits are reached for swaptions with very short maturity of 1 month which seems natural given the reconstruction behavior of the model shown in Figure \ref{VolDiff}. This is in particular in congruence with mean absolute errors for SABR fits calibrated on the observed volatility which exhibit largest misfits in the high tenor - low maturity range as well.  For swaptions with maturity between 6 months and 2 years and tenor between 1 and 5 years the best fit is obtained. A similar goodness-of-fit behavior can be found when averaging the mean absolute errors shown here over all samples in the test set with a maximum MAE of 2.76 obtained for the 1 month - 10 year contract. To put this into perspective, a difference of 2.76 basis points in Bachelier implied volatility makes up a 7.37\% price difference for the at-the-money swaption contract.

Next to the direct evaluation of the volatility fits, other evaluation metrics can be applied when judging the quality of a SABR fit that is applied to a Pseudo-Gibbs imputed volatility smile. In particular, the delta hedging performance of the approach is analyzed in the following subsection. 

\vspace*{0.5cm}
\subsection{Delta hedging using VAE-imputed volatility data}
In this section we apply the Pseudo Gibbs imputation algorithm to the context of delta hedging a synthetically simulated European swaption contract maturing 1 year in the future with a tenor of 1 year ($T_0 = 1$, $T_M = T_0 + 1$). Notational prerequisites are summarized briefly in Appendices A and B. The precise methodology for the hedging study is as follows:
\begin{itemize}
\item[1)] On the $(T_0, T_M)$-grid described in Section 3 we fix $\alpha_{T_0, T_M}, \nu_{T_0, T_M}$ and $\rho_{T_0, T_M}$ parameters, which where calibrated to the swaption cube on 20 Oct 2021. $\beta$ is fixed to a value of 0.5.
\item[2)] For each maturity-tenor combination we generate a 1 year timeseries of forward swap rates $F_t$ by a discretization of the SABR stochastic differential equation. In order to obtain realistic dependencies between forward swap rates of different maturities and tenors, we use the same normally distributed increments of the Brownian motion processes for trajectory generation on the whole grid. The swaption corresponding to the simulated timeseries for a maturity and a tenor of 1 year is the one we want to hedge.
\item[3)] At each point in time we reconstruct a theoretical swaption volatility cube from the simulated paths by interpolating the forward swap rates on the $(T_0, T_M)$-grid and by applying \eqref{normvol}. Afterwards, we mask 70\% of the points on the swaption volatility cube at each point in time. This is the volatility cube the practitioner is supposed to observe in the market.
\item[4)] We reconstruct the full cube using the trained VAE-imputation model at each point in time. Afterwards the values $\alpha_{T_0, T_M}$, $\nu_{T_0, T_M}$ and $\rho_{T_0, T_M}$ are calibrated from the reconstructed cube on the $(T_0, T_M)$-grid. In order to obtain SABR parameters corresponding to the maturity of the contract we want to hedge, we interpolate the obtained SABR parameters on the $(T_0, T_M)$-grid once again to the present maturity of the original 1 year swaption.
\item[5)] Using the SABR parameters obtained in step 4) at each point in time, we calculate the delta of the swaption that we want to hedge. Using a forward swap with the same maturity and tenor, we obtain a dynamic delta-neutral portfolio as described in Appendix B.
\end{itemize}

Figure \ref{Delta_Fig} examines the hedging performance of the described approach along a specific simulated path of the 1 year - 1 year forward swap rate. We compare the performance of the delta-neutral portfolio obtained by the VAE-imputation algorithm and by interpolation of the cube with missing values to the performance of the theoretical delta-neutral portfolio that could be obtained if the practitioner hedging the swaption had perfect knowledge in advance about the risk-neutral parameters $\alpha$, $\beta$, $\nu$ and $\rho$ generating the observed path. Analogously to the results of Section 4.1, this demonstrates the superiority of the VAE imputation approach over interpolation of missing values. The swaption's notional was set to $N = 100\,000$ and payment dates where set to a quarterly tenor after maturity at $T_0=1$. For simplicity, we assume 360 trading days per year without weekend effects. Moreover, for the ease of exposition, a deterministic exponential discount rate structure was presumed for the zero-coupon bond values $P(t, T_i)$. 

As it was briefly discussed in Appendix B, note that the construction of a delta-neutral portfolio described here does not lead to a perfect replication of a swaption contract by a self-financing portfolio as it is possible in complete stochastic models. Neverless, the approach of the current section yields a dynamic assessment of the hedging performance and  is comparable with \textcite{RPW}.

In order to track the performance of the described approach along multiple trajectories we measure the root mean squared error (RMSE) between final cumulated predicted price differences and the theoretical final option prices (see the lower part of Figure \ref{Delta_Fig}) along $10\,000$ simulated paths of the SABR dynamics. The results are shown in Table \ref{Errors} below as percentages of the swaption's notional value of $100\,000$. One can observe that the differences between the theoretical delta-neutral portfolio and the VAE-imputation portfolio become negligible when using larger rebalancing periods. However, VAE-imputation errors exhibit a much slower decline when decreasing the rebalancing period compared to the theoretical delta-neutral portfolio.

\begin{table}[H]
\centering
\begin{tabular}{|C{3cm}|C{4.5cm}|C{4.5cm}|}
\hline
\large{\textbf{Rebalancing}} & \large{\textbf{Theoretical Portfolio}} & \large{\textbf{Imputation Portfolio}} \\\hline
1 week & $2.059 \cdot 10^{-2}$ \%  & $3.609 \cdot 10^{-2}$ \% \\\hline
1 day &  $1.350 \cdot 10^{-2}$ \% & $2.300 \cdot 10^{-2}$ \% \\\hline
1 hour & $0.569  \cdot 10^{-2}$ \% & $1.482 \cdot 10^{-2}$ \% \\\hline
1 minute & $0.155 \cdot 10^{-2}$ \% & $1.293 \cdot 10^{-2}$ \% \\\hline
\end{tabular}
\caption{RMSE between cumulated predicted price differences and theoretical final swaption prices as a percentage of the swaption's notional of $100\,000$ for  rebalancing periods between 1 week and 1 minute using the theoretical delta-neutral and the VAE-imputation portfolio.}
\label{Errors}
\end{table}

\begin{figure}[H]
\centering
\includegraphics[width=\textwidth]{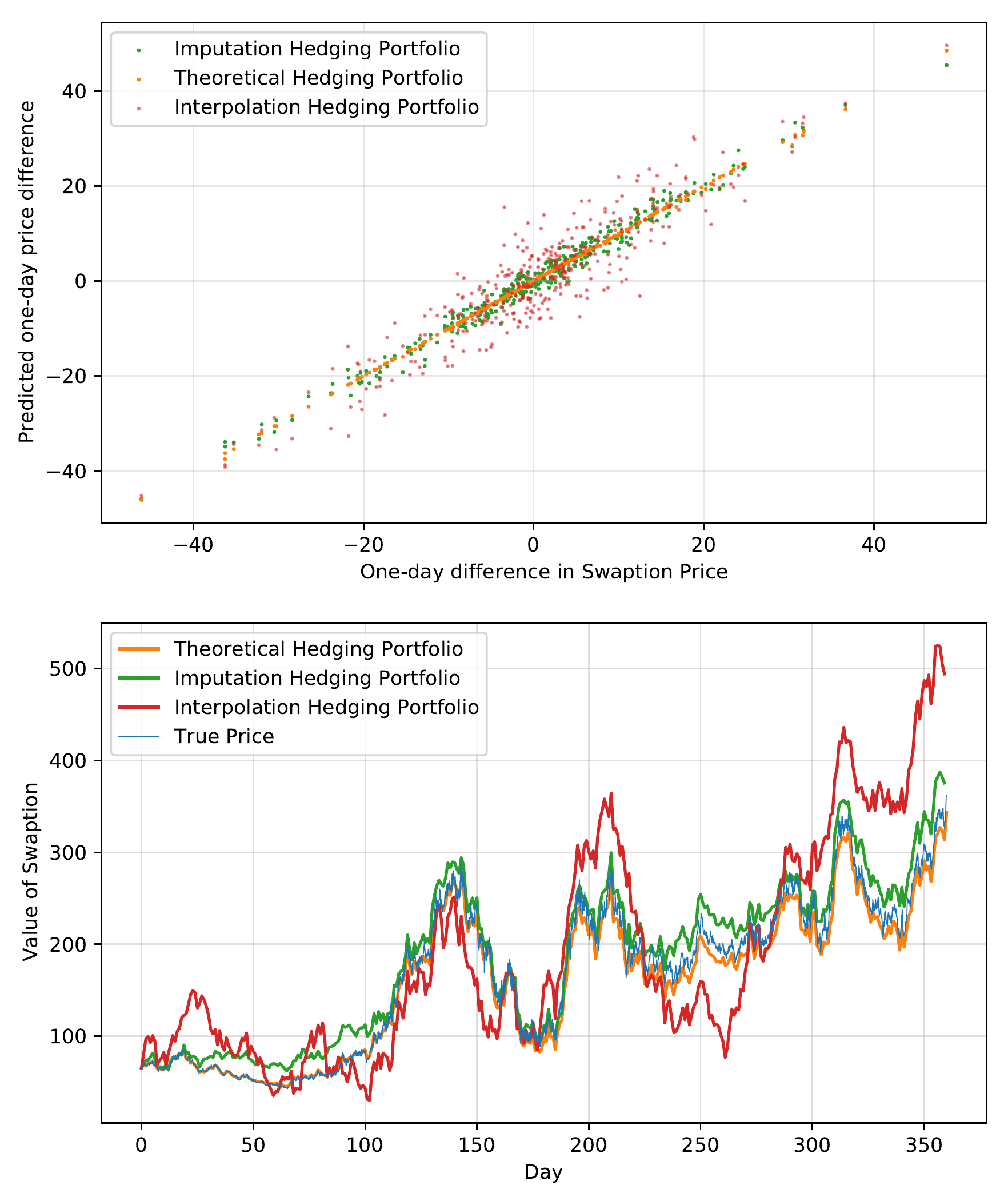}
\caption{Upper Figure: Actual one-day changes in swaption price vs. one-day changes in swaption price predicted by the delta-neutral portfolios using the theoretical hedging portfolio as well as the hedging portfolios obtained from the VAE imputation and interpolation approaches compared in Figure \ref{SABR_Reco}. The $R^2$ value obtained from a regression of the predicted changes on the actual changes is 0.9778 for the VAE-imputation portfolio, 0.8532 for the interpolation portfolio and 0.9998 for the theoretical hedging portfolio. Lower Figure: The actually realized swaption price was plotted next to the cumulated predicted price differences from the upper figure.}
\label{Delta_Fig}
\end{figure}

\section{Extensions and further research possibilities}
In the preceding sections we demonstrated how the geometry of an implied volatility cube containing missing values can be inferred by learning stochastic latent representations via variational autoencoders in an approximate Gibbs sampling environment. The imputed estimates of missing quotes were afterwards used to fit the SABR stochastic model on volatilty smiles. 

There are plenty of possibilities to extend the basic imputation algorithm presented in the second section. First, \textcite{MF1} propose to make use of a Metropolis-within-Gibbs extension of the basic Pseudo-Gibbs imputation of \textcite{RMW} in order to asymptotically sample from the true conditional distribution $p(x_{\text{miss}} | x_{\text{obs}})$ instead of the variational approximation $q(x_{\text{miss}} | x_{\text{obs}})$. Here, instead of sampling $z^{(t+1)}$ from $q_\theta(z|(x_{\text{obs}}, x_{\text{miss}}))$ in step 3) of the algorithm in Section 2, we insert a Metropolis-Hastings step in the algorithm which makes sampling from the true posterior asymptotically tractable.  This procedure, however, requires very thorough tuning of the Metropolis-Hastings steps due to the very high dimensionality (4998-dimensional volatility data) of the problem. Apart from that, there is a number of variance reduction techniques that can be applied to the basic Gibbs sampling procedure, for example a basic Rao-Blackwellization procedure. 

Second, in order to better capture the highly nonlinear dependence between the components of volatility cubes, other decoder architectures for the pretrained variational autoencoder could prove to be fruitful. For example, convolutional variational autoencoders as well as a different parameterization of the Gaussian covariance matrix as a rank-1 matrix with diagonal correction would be possible. \textcite[Section 4.3]{RMW} propose the parameterization of the precision $\Sigma^{-1} = D + uu^\top$ where $D$ is diagonal and $u$ is a vector, which allows for arbitrary rotations of the Gaussian distribution along one principle direction with relatively few additional parameters (see \textcite{MP}).

The algorithm presented in Section 2 utilizes synthetically generated training data samples, as obtained in Section 3.  Alternatively, robustness of the Pseudo-Gibbs imputation approach can be studied when the VAE model is trained on samples that already include missing values. \textcite{MF2} propose a method based on the importance-weighted autoencoder of \textcite{BGS} to train deep latent variable models on training data including missing values. Finally, in order to improve the predictive power of the variational decoder model, we could also apply a recurrent network structure, accounting for time dependencies of surfaces or cubes because, after all, even if the observed values of today's volatility cube are predicted to come from a cube of a particular shape, the model should predict today's volatility cube to not differ very much of yesterday's cube in shape.

\section*{References}
\nocite{*}
\printbibliography[heading = none]

\newpage
\begin{appendices}
\section{The SABR Stochastic Volatility Model}
The SABR stochastic volatility model of \textcite{HetAl} has become one of the effective standards in interest rate derivatives modeling which stems from its ability to accurately fit market implied volatility smiles with only four parameters $\alpha$, $\rho$, $\nu$ and $\beta$ to fit. Here, the parameter $\beta$ controls the overall smile dynamic and is usually fitted ex ante from historical series analysis for the relevant market. 

In our studies, we fit the parameters of a negative rate adjusted \textit{shifted SABR model} to the Bachelier model implied normal volatility using the serial expansion formulas for the implied volatility from \textcite{HetAl} with slight corrections from \textcite{O} and using the explicit initial guesses derived in \textcite{LK} for the iterative calibration procedure. In our given context, the shifted SABR model describes the dynamics of the forward swap rate $F$ for a given swaption maturity $T_0$ and tenor $T_M$ under the forward swap measure\footnote{The forward swap measure $Q^{T_0, T_M}$ is the equivalent martingale measure associated with the numeraire process $N_t := \sum_{i=1}^m \delta_i P(t, T_i)$, see the notation on the following page as well as \textcite{BM}.} $Q^{T_0, T_M}$ with payment dates $T_1, \dots, T_M$ by the two-factor CEV-type stochastic differential equations
\begin{align*}
\dd F_t &= \sigma_t (F_t + b)^\beta \dd W_t^{(1)} \\
\dd \sigma_t &= \nu \sigma_t \dd W_t^{(2)}\\
F_0 &= f \\
\sigma_0 &= \alpha
\end{align*}
where $f$ is the current forward swap rate, $b$ is a displacement parameter allowing for negative rates and where $W^{(1)}$ and $W^{(2)}$ are two standard Brownian motions with correlation $\rho$. 

In practice, the value of a European swaption is generally quoted in terms of its Bachelier- or Black-model implied volatility. We refer to Bachelier-model implied volatility, which is typically quoted in basis points, as normal volatility or basis point volatility while we refer to Black-model implied volatility, which is typically quoted in percentage, as lognormal volatility. Given the low-interest regime that dominates current markets, we opt to represent swaption values in terms of normal volatility $\sigma_N$. Let $K$ denote the strike of the swaption, i.e. the fixed interest rate underlying the swap that the can be entered by the holder of the swaption at time $T_0$. Following the asymptotic expansion of \textcite{HetAl} and the simplifications in \textcite{LK} we have the approximate formula for $F_t \neq K$ and $\beta \in [0, 1]$
\begingroup\makeatletter\def\f@size{8.5}\check@mathfonts
\begin{equation}
\label{normvol}
\sigma_N(t) = \sigma_N(F_t, K, T, \alpha, \beta, \nu, \rho) \approx \frac{F_t - K}{\hat{x}(\zeta)}\left[1 + \left(g + \frac{1}{4}\rho \nu\alpha\beta(F_t + b)^{\frac{\beta - 1}{2}}(K + b)^{\frac{\beta - 1}{2}} + \frac{1}{24}(2 - 3 \rho^2)\nu^2\right)(T_0 - t)\right]
\end{equation} 
\endgroup
where, for $\beta \in [0, 1]$, the values $g$, $\zeta$ and $\hat{x}(\zeta)$ are given by
\begin{align*}
g &= \frac{1}{24}(\beta^2 - 2\beta)(F_t + b)^{\beta - 1}(K + b)^{\beta - 1} \alpha^2 \\
\zeta &= \frac{\nu}{\alpha(1 - \beta)}\left((F_t + b)^{1- \beta} - (K + b)^{1 - \beta}\right)  \qquad (\beta \neq 1) \\
\hat{x}(\zeta) &= \frac{1}{\nu}\log\left(\frac{\sqrt{1 - 2\rho\zeta + \zeta^2} - \rho + \zeta}{1 - \rho}\right).
\end{align*}
and where $\zeta = \frac{\nu}{\alpha}\log\left(\frac{F_t + b}{K + b}\right)$  if $\beta = 1$. If $F_t = K$ the normal volatility is given by
\begingroup\makeatletter\def\f@size{9.5}\check@mathfonts
\begin{equation}
\label{normvol0}
\sigma_N(t) = \sigma_N(F_t, K, T, \alpha, \beta, \nu, \rho) \approx \alpha(F_t + b)^\beta \left[1 + \left( g + \frac{1}{4}\rho\nu\alpha\beta(F_t + b)^{\beta - 1} + \frac{1}{24}(2 - 3\rho^2)\nu^2\right)(T_0-t)\right].
\end{equation} 
\endgroup

\newpage
\noindent In the Bachelier model, the forward swap rate under the forward swap measure is modeled by
\begin{align*}
\mathrm{d}F_t &= \sigma_N \dd W_t \\
F_0 &= f
\end{align*} 
which possesses the solution $F_t = f + \sigma_N W_t$, i.e. a Brownian motion with scale and drift. Having obtained the implied normal volatility in the SABR model from \eqref{normvol} or \eqref{normvol0} respectively, one can easily obtain the SABR swaption price via the Bachelier model valuation formulas (see e.g. \textcite{CWL}), i.e. for $t \in [0, T_0)$
\begin{equation}\label{price}
V^{\mathrm{Bachelier}}_t = N \cdot \left[\sum_{i=1}^{m} \delta_i P(t, T_i)\right] \sigma_N \sqrt{T_0 - t} \big(d [\Phi(d) - R] + \varphi(d)\big), 
\end{equation}
where $N$ denotes the notional amount of the swaption, $\delta_i$ is a fraction denoting the day-count convention for the period $[T_{i-1}, T_i]$, $P(t, T_i)$ denotes the discount factor for the period $[t, T_i]$ usually measured by the price of an according zero-coupon bond, $d = \frac{F_t - K}{\sigma_N\sqrt{T_0 - t}}$ and $\Phi$ and $\varphi$ denote the cumulative standard normal distribution function and the standard normal density respectively. The term $\sum_{i=1}^{m} \delta_i P(t, T_i)$ is commonly referred to as the present value of a basis point. The value of $R$ is set to 0 if the swaption is a payer swaption (i.e. the holder of the swaption pays the fixed leg) and set to $1$ if the swaption is a receiver swaption (i.e. the holder of the swaption receives the fixed leg). When computing the SABR price of a swaption, \eqref{price} is used in conjuction with $\sigma_N(t)$ from \eqref{normvol} or \eqref{normvol0}.

\section{Delta hedging in the SABR model}

The idea of basic dynamic delta hedging a short position in a swaption consists in taking a long position in the forward swap contract corresponding to the swaptions underlying payment structure. The size of this long position at time $t$ will be denoted $m_t$, while the value $\Delta_t = \frac{\partial V^{\mathrm{SABR}}_t}{\partial F_t}$ will be called the delta of the swaption where differentiation takes place with respect to the value function in the SABR model. In order to make the combined portfolio of those two positions independent of fluctuations in the forward swap rate $F_t$, $m_t$ has to fulfill the condition
\begin{equation}\label{Delt}
m_t \frac{\partial V^{\mathrm{Swap}}_t}{\partial F_t} - \Delta_t = 0,
\end{equation} 
where $V^{\mathrm{Swap}}_t$ denotes the value of the forward swap contract in the portfolio. In order to obtain $\Delta_t$ in the SABR model, we can decompose
\[\Delta_t = \frac{\partial V^{\mathrm{SABR}}_t}{\partial F_t} = \frac{\partial V^{\mathrm{Bachelier}}_{t}}{\partial F_t} + \frac{\partial \sigma_N(t)}{\partial F_t} \frac{\partial V_t^{\mathrm{Bachelier}}}{\partial \sigma_N},\]
where $\partial V^{\mathrm{Bachelier}}_{t} \,/ \, \partial F_t$ denotes the delta of the swaption in the Bachelier model, i.e. the partial derivative of \eqref{price} with respect to $F_t$ and where $\partial V^{\mathrm{Bachelier}}_{t} \,/ \, \partial \sigma_N$ denotes the Vega of the swaption in the Bachelier model, i.e. the partial derivative of \eqref{price} with respect to $\sigma_N$. Plugging in the corresponding partial derivatives calculated from \eqref{price} and \eqref{normvol} we obtain for $\beta \in (0, 1)$ and $F_t \neq K$
\begin{equation}\label{Delta}
\Delta_t = N \cdot \left[\sum_{i=1}^{m} \delta_i P(t, T_i)\right]\left[\Phi(d) + \sqrt{T_0 - t}\varphi(d) \big(\sigma_N(t) \cdot \kappa + \tau\big) - R\right],
\end{equation} 
where
\begin{align*}
\tau &= \frac{F_t - K}{\hat{x}(\zeta)(F_t + b)} (\beta - 1) \left[ g + \frac{1}{8}\rho \nu \alpha \beta (F_t + b)^{\frac{\beta - 1}{2}} (K + b)^{\frac{\beta - 1}{2}}\right] (T_0 - t) \\
\kappa &= (F_t - K)^{-1} - \frac{(F_t + b)^{-\beta}}{\alpha \hat{x}(\zeta) \sqrt{1 - 2\rho \zeta + \zeta^2}}
\end{align*}
if $F_t \neq K$ whereas, for $F_t = K$
\begingroup\makeatletter\def\f@size{10}\check@mathfonts
\begin{align*}
\tau = \alpha (\beta - 1) (F_t + b)^{\beta - 1} \left[ g + \frac{1}{8}\rho \nu \alpha \beta (F_t + b)^{\beta - 1}\right] (T_0 - t), \qquad \kappa = \frac{1}{2}\left[ \beta (F_t + b)^{-1} - \frac{\rho \nu}{\alpha}(F_t + b)^{-\beta}\right].
\end{align*}
\endgroup
\textcite{B} derives an alternative formula for delta in the SABR model that accounts for a forward swap rate change induced jump in instantaneous volatility. For simplicity, we will not focus on Bartlett's delta here. Using the formulas \eqref{Delt} and \eqref{Delta}, the corresponding long position needed in the forward swap for delta neutrality is easily obtained using that $V^{\mathrm{Swap}}_t$ is given by
\[V^{\mathrm{Swap}}_t = (1 - 2R) N \cdot \left[\sum_{i=1}^{m} \delta_i P(t, T_i)\right] (F_t - K).\]

Other methods of hedging a swaption include for example the use of a portfolio of zero coupon bonds, see e.g. \textcite{DSB}. Note that the constructed delta-neutral portfolio approach described here does not replicate the swaption contract perfectly like it is the case e.g. in the plain-vanilla Black model. Nevertheless, examining the differences between actual swaption price changes and changes in the dynamic portfolio value between two rebalancing dates, as it is done e.g. by \textcite{RPW}, yields an assessment of the hedging performance of a particular stochastic model like it was done in Section 4.2.

\section{Model architecture and considered market cubes}
\begin{figure}[H]
\centering
\includegraphics[trim = 0mm 0mm 0mm 5mm, clip, width=0.88\textwidth]{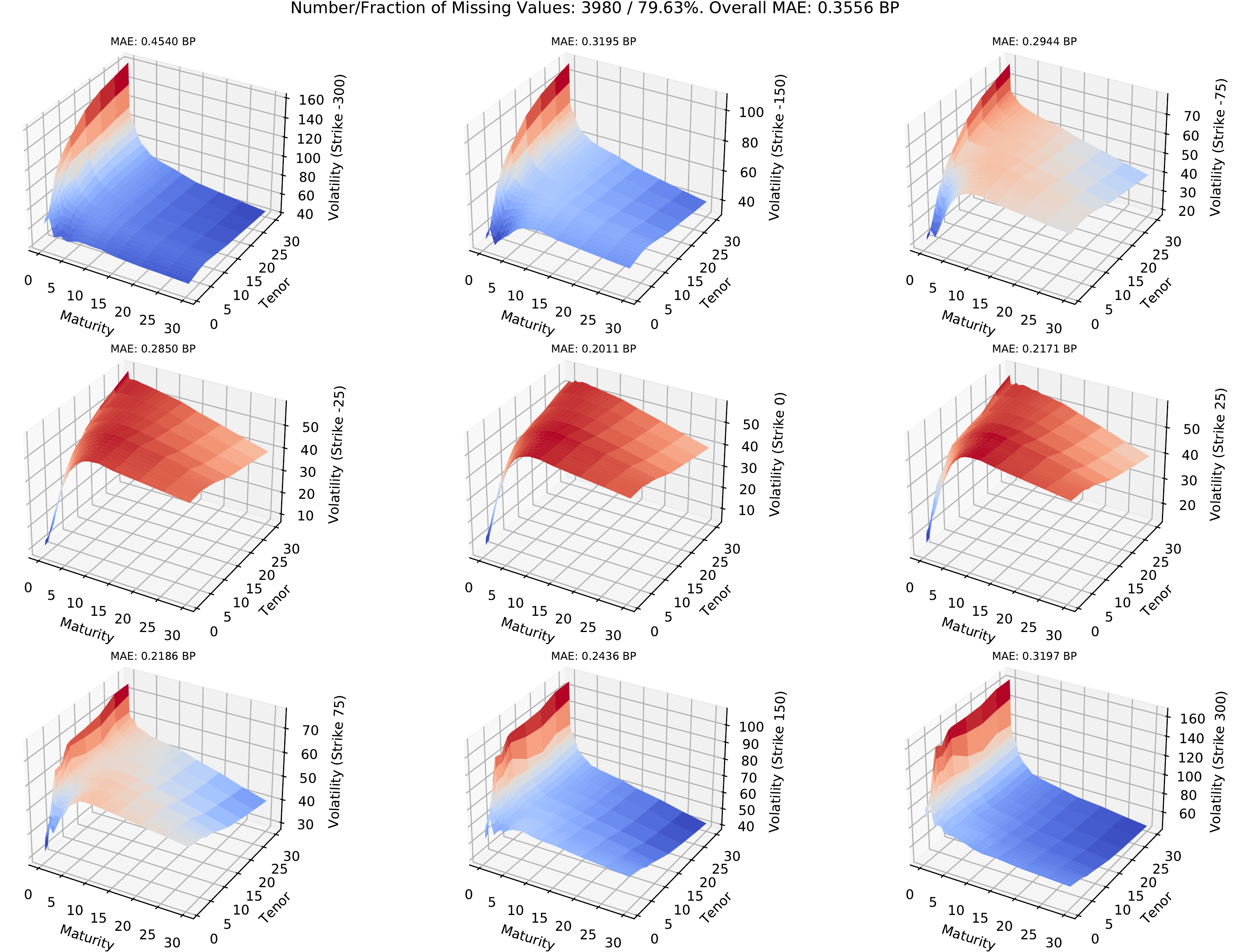}
\caption{Approximately 79.6\% of the volatility quotes of the cube on 20 October 2021 were masked and treated like missing data. The cube depicted is the obtained cube after missing data imputation with the Pseudo-Gibbs sampling approach described above.}
\label{Imputations20}
\end{figure}

\begin{figure}[H]
\centering
\includegraphics[width=0.92\textwidth]{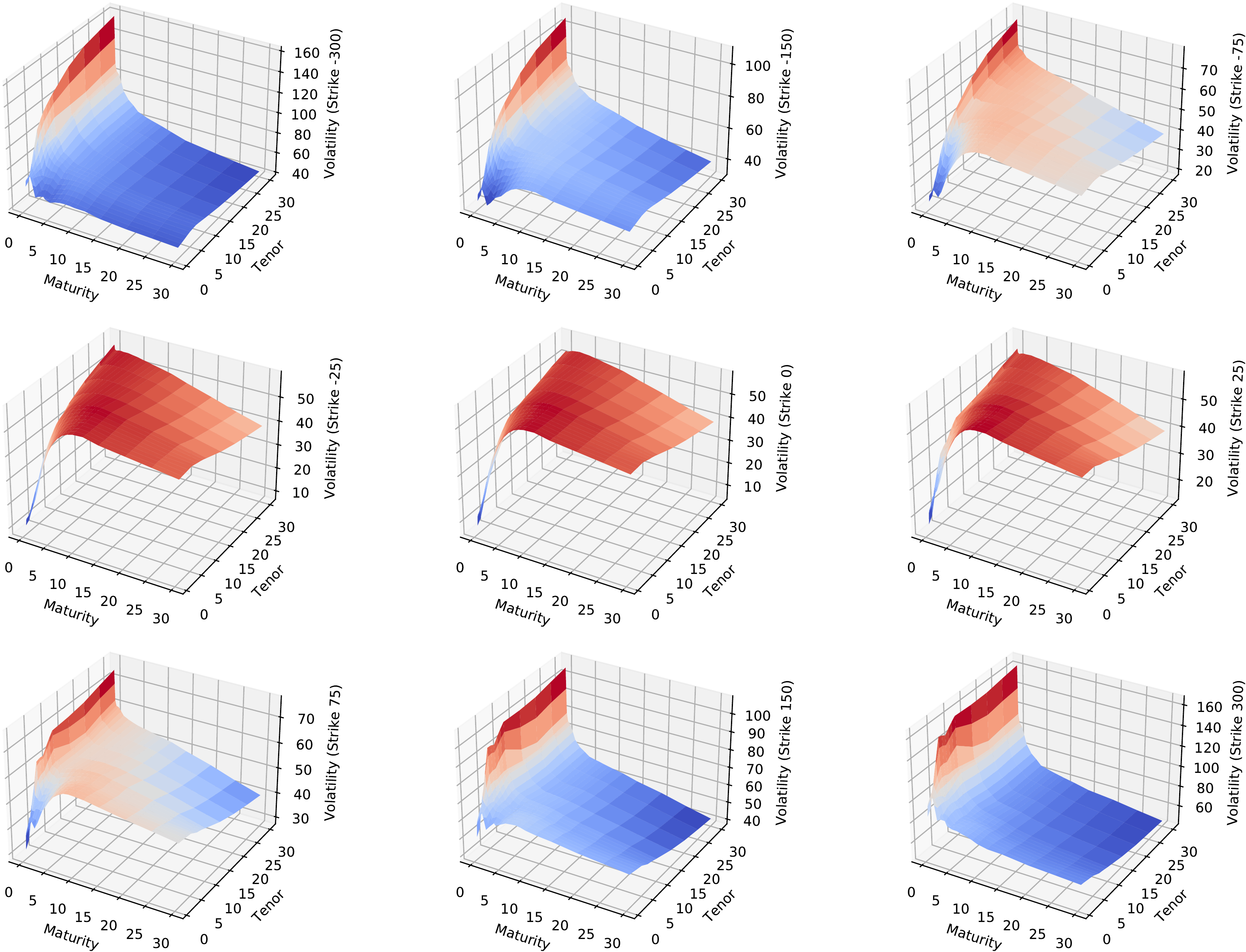}
\caption{Market observed volatility cube on 20 October 2021.}
\label{TrueCube20}
\end{figure}

Figures \ref{Imputations20} and \ref{TrueCube20} show the considered market swaption volatility cube observed on 20 October 2021 and its Pseudo-Gibbs reconstruction, the differences of which where shown in Figure \ref{VolDiff}.

\begin{figure}[H]
\centering
\includegraphics[width=0.85\textwidth, height=7.7cm]{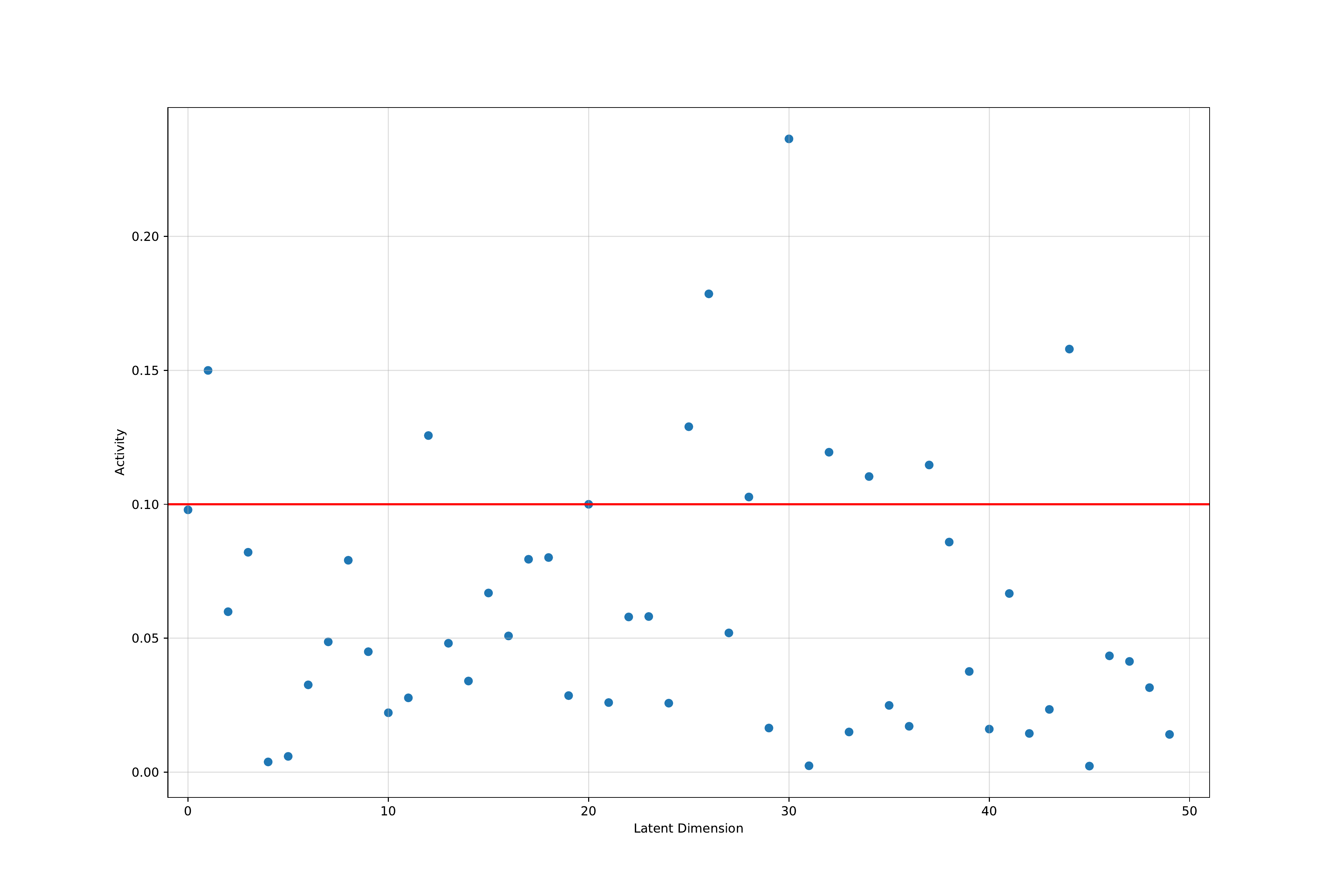}
\caption{Latent Activity Statistics for a VAE model with 50 bottleneck units. Approximately 11 of the 50 units where active in the sense described above, motivating the employed VAE architecture with 10 hidden units.}
\label{Activity}
\end{figure}

The trained VAE inference model used throughout the paper was a standard variational autoencoder with a ten-dimensional latent space which was trained for 50000 epochs on 10000 synthetically generated swaption cubes by the procedure described in Section 2. The choice of dimensionality for the bottleneck layer was based on a latent activity statistic proposed by \textcite{BGS}: The activity of each latent node is measured by $$A_u := \mathrm{Var}_x\big(\E_{q_\theta(z|x)}(z)\big)$$ and we call a node inactive if $A_u < 0.1$\footnote{In their paper, \textcite{BGS} chose an activity threshold of 0.01.}. Figure \ref{Activity} shows $A_u$ for a trained VAE model with 50 latent units. One can see that of the 50 units approximately 11 units are active which motivates our VAE architecture with 10 hidden units. Using this architecture, all units remain active.

Both the encoder and decoder submodel of the VAE were equipped with four hidden layers with 250, 200, 150 and 100 units respectively using ReLU activations. The layer weights were initialized normally distributed with a variance of 1/30. In our experiments we found that fine tuning the kernel initializer variance had quite a large impact on the stability of training on synthetic data. The employed optimizer was the Adam algorithm with a learning rate of $10^{-6}$.


\end{appendices}
\end{document}